\documentclass[runningheads]{llncs}
\usepackage[T1]{fontenc}
\usepackage{graphicx,verbatim, multirow, multicol,booktabs, rotating}
\usepackage{amsfonts, amsmath}
\usepackage{glossaries}
\usepackage{pifont}
\usepackage{bbold}
\usepackage{hyperref}
\usepackage{color}

\begin{document}
\title{CyclePose - Leveraging Cycle-Consistency for Annotation-Free Nuclei Segmentation in Fluorescence Microscopy}
\titlerunning{CyclePose - Annotation-Free Nuclei Segmentation}
\author{Jonas Utz\inst{1} \and
Stefan Vocht\inst{1} \and
Anne Tjorven Büssen\inst{1} \and
Dennis Possart\inst{1} \and
Fabian Wagner\inst{2} \and
Mareike Thies\inst{1,2} \and
Mingxuan Gu\inst{2} \and
Stefan Uderhardt\inst{3} \and
Katharina Breininger\inst{4}}
\authorrunning{J. Utz et al.}
\institute{Department Artificial Intelligence in Biomedical Engineering, Friedrich-Alexander-Universität (FAU), Erlangen, Germany \\
\email{jonas.utz@fau.de}\\
\and
Pattern Recognition Lab, FAU, Erlangen, Germany\\
\and
Department of Internal Medicine 3 and Institute for Clinical Immunology, FAU, Erlangen, Germany\\
\and
Center for AI and Data Science (CAIDAS), Universität Würzburg, Würzburg, Germany
}

\maketitle              %

\newacronym{ANN}{ANN}{Artificial Neural Network}
\newacronym{AI}{AI}{Artificial Intelligence}
\newacronym{ML}{ML}{Machine Learning}
\newacronym{MLP}{MLP}{Multilayer Perceptron}
\newacronym{DNN}{DNN}{Deep Neural Network}

\newacronym{MSE}{MSE}{Mean Squared Error}
\newacronym{MAE}{MAE}{Mean Absolute Error}

\newacronym{SGD}{SGD}{Stochastic Gradient Descent}

\newacronym{ReLU}{ReLU}{Rectified Linear Unit}

\newacronym{GAN}{GAN}{Generative Adversarial Network}
\newacronym{cGAN}{cGAN}{Conditional Generative Adversarial Net}
\newacronym{CycleGAN}{CycleGAN}{Cycle-Consistent Generative Adversarial Network}
\newacronym{DCGAN}{DCGAN}{Deep Convolutional Generative Adversarial Network}
\newacronym{LSGAN}{LSGAN}{Least Squares Generative Adversarial Network}

\newacronym{DCT}{DCT}{discrete cosine transform}

\newacronym{NMS}{NMS}{Non-Maximum Suppression}

\newacronym{IoU}{IoU}{Intersection over Union}
\newacronym{TP}{TP}{True Positive}
\newacronym{FP}{FP}{False Positive}
\newacronym{FN}{FN}{False Negative}
\newacronym{TN}{TN}{True Negative}

\newacronym{AP}{AP}{Average Precision}

\newacronym{CT}{CT}{computed tomography}

\newacronym{FAU}{FAU}{Friedrich-Alexander-Universität Erlangen-Nürnberg}

\newacronym{WYGIWYAF}{WYGIWYAF}{What You Get Is What You Ask For}
\newacronym{WYSIWYG}{WYSIWYG}{What You See Is What You Get}

\begin{abstract}
In recent years, numerous neural network architectures specifically designed for the instance segmentation of nuclei in microscopic images have been released. These models embed nuclei-specific priors to outperform generic architectures like U-Nets; however, they require large annotated datasets, which are often not available. Generative models (GANs, diffusion models) have been used to compensate for this by synthesizing training data. These two-stage approaches are computationally expensive, as first a generative model and then a segmentation model has to be trained. We propose CyclePose, a hybrid framework integrating synthetic data generation and segmentation training. CyclePose builds on a CycleGAN architecture, which allows unpaired translation between microscopy images and segmentation masks. We embed a segmentation model into CycleGAN and leverage a cycle consistency loss for self-supervision. Without annotated data, CyclePose outperforms other weakly or unsupervised methods on two public datasets. Code is available at \url{https://github.com/jonasutz/CyclePose}  


\end{abstract}
\section{Introduction}
 Nuclei segmentation is essential in quantitative cell biology and digital pathology, enabling downstream tasks such as cell counting, tracking, and morphological analysis~\cite{basu_survey_2023}.  As a key structural component, the nucleus is often used to identify individual cells, especially when cell boundaries are unclear. Accurate nuclei segmentation remains challenging owing to variations in imaging conditions, including signal-to-noise ratio, staining intensity, and microscope settings. Densely packed nuclei often have touching or overlapping boundaries, which makes instance separation difficult. While early approaches relied on intensity thresholding and linear filtering~\cite{50_years_cell_segmentation}, deep learning now dominates because of its ability to extract complex features and contextual information. Several specialized architectures have been developed for cell and nuclei segmentation. Notable models include StarDist~\cite{schmidt2018} and Cellpose~\cite{stringer_cellpose_2021}, which incorporate domain-specific priors, such as shape and distribution of nuclei, to outperform generic instance segmentation methods such as Mask R-CNN~\cite{He2017} and U-Net~\cite{ronneberger_u-net_2015}. However, these models rely on supervised learning and require large, annotated datasets. Nuclei annotation is particularly labor-intensive and requires expert knowledge and training to delineate individual instances consistently. This creates a bottleneck for training deep learning models at scale, necessitating methods that require little or no annotated data. A promising direction is the use of generative models such as CycleGAN to synthesize realistic microscopy images from masks. Conventionally, synthetic data generated through such models can be used to train a segmentation network; however, this two-step approach is computationally expensive and may introduce a domain gap between synthetic and real images. 
 
 \noindent In this study, we tackle these limitations and propose an annotation-free approach that directly integrates a Cellpose-based segmentation model into a CycleGAN-inspired framework. By leveraging the cycle consistency loss, we train the segmentation network without requiring explicitly annotated training data, making our approach effectively unsupervised. This hybrid architecture offers reduced parameter count, faster training times, and allows the segmentation model to interact directly with real microscopy images during training. By avoiding separately trained synthetic data models, our approach elegantly streamlines nuclei segmentation and improves the adaptation to real-world imaging conditions.
 
\section{Related Work}
Deep learning-based models have largely replaced traditional segmentation methods (e.g., thresholding, linear filtering, and watershed transform) due to their superior performance in cell and nuclei segmentation. Several algorithms have been developed specifically for this task. \textit{StarDist}~\cite{schmidt2018} models objects as star-convex polygons, leveraging prior knowledge that nuclei are typically blob-like to predict distances from a central point and define nuclear boundaries. This representation mitigates issues such as cell merging in per-pixel segmentation and inaccuracies in bounding-box-based approaches. \textit{Cellpose}~\cite{stringer_cellpose_2021} predicts a flow field for each image pixel, guiding pixels toward their respective nuclei centers. Pixels that converge to the same point are grouped into an instance. This approach improves segmentation in nuclei-dense regions, where per-pixel classification or bounding-box-based methods often fail. However, both \textit{StarDist} and \textit{Cellpose} rely on annotated training data which limits their utility since manual annotation is a labor-intensive and time-consuming process. To circumvent this, various unsupervised nuclei segmentation approaches have been proposed. Many unsupervised nuclei segmentation methods rely on a two-step process where, first, synthetic image-mask pairs are created, and second, these are used to train a segmentation model. A widely used strategy for generating synthetic masks involves deforming ellipses or leveraging masks from a different domain~\cite{brieu2019domain}, as demonstrated in prior studies~\cite{boehland_2019,utz2023focuscontentnoiseimproving,Hoblurpoisson,Spcyclegan,Nisnet3d}. Generative adversarial networks (GANs), such as CycleGAN, can then be employed to synthesize corresponding images, creating a paired dataset for training~\cite{Spcyclegan,utz2023focuscontentnoiseimproving,brieu2019domain}. Alternatively, nuclei textures can be synthesized using Perlin noise~\cite{Perlin}, or synthetic images can be obtained through simulation-based methods such as \textit{CytoPacq}~\cite{Cytopaq}.
Although unsupervised methods eliminate the need for annotations, semi-supervised approaches can reduce annotation effort while still benefiting from annotated data. \textit{DenoiSeg}~\cite{buchholz_denoiseg_2020} requires only a small set of annotated images by integrating denoising and segmentation into a single network. Denoising simplifies the segmentation task, and by combining both tasks within a single model, a co-learning effect emerges, where the simultaneous learning of both tasks mutually reinforces performance.

\noindent Unlike prior unsupervised approaches that rely on synthetic image-mask pairs for training, our method directly integrates segmentation within a CycleGAN-inspired framework, eliminating the need for separate synthetic data generation. This reduces computational overhead and minimizes the domain gap, enabling more robust adaptation to real microscopy images.
\section{Methods}
\label{sec:methods}
\begin{figure}[t]
    \centering
    \includegraphics[width=\textwidth]{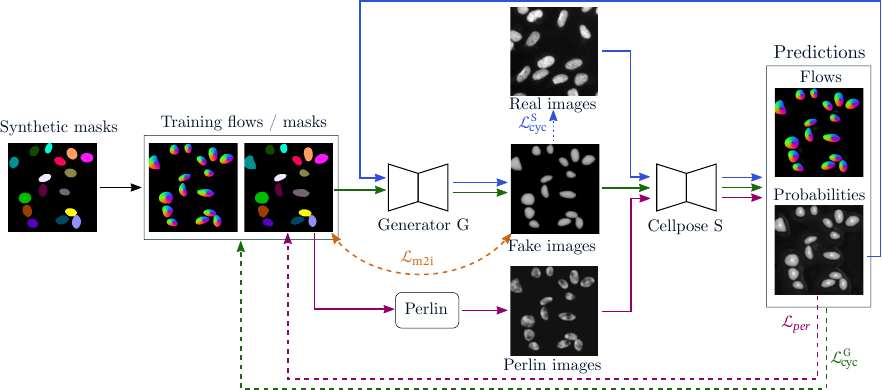}
    \caption{Overview of the cycles in CyclePose: Synthetic masks from rotated ellipses undergo elastic deformation to create training masks. Adding Perlin noise-based texture, Gaussian blur, and Poisson noise produces Perlin images, which serve as auxiliary inputs for Cellpose alongside fake and real images. Discriminators are omitted for brevity. During inference only $S$ is used.}
    \label{fig:main}
\end{figure}
Our proposed method, \textbf{CyclePose}, is an unsupervised nuclei segmentation framework that integrates the gradient flow representation of Cellpose within a CycleGAN architecture. The method relies on two key assumptions: (1) synthetic segmentation masks for nuclei images can be easily generated using deformed ellipses, and (2) the cycle consistency loss of a CycleGAN can serve as an effective proxy for supervision in the absence of paired data.

CyclePose consists of four primary network components: an image generator $G$, a segmentation model $S$ (based on Cellpose), and two domain-specific discriminators $D_{img}, D_{seg}$. 
$G$ translates synthetic segmentation masks into microscopy images, while the 
$S$ takes microscopy images and predicts the horizontal and vertical gradients of the Cellpose flows and a probability map for nuclei segmentation. Synthetic masks for $G$ can be obtained by applying random elastic deformations to binary masks of ellipses, while $S$ is trained on real microscopy images. Following the CycleGAN paradigm, the network outputs are passed back through the counterpart networks, forming a cycle (Fig.~\ref{fig:main}). This resembles a self-supervised training of $S$ on generated microscopy images using the synthetic masks as ground truth. Hence, our cycle loss $\mathcal{L}_{cyc}$ is a weighted sum of an L2 loss for the flows and a cross-entropy loss for the probabilities as described in~\cite{stringer_cellpose_2021}. For image reconstruction, we use an L1 loss.
To enable an annotation-free training pipeline, we further introduce the following two loss components.

\noindent\textbf{Perlin Loss} CycleGANs are known for unstable training, which we address by introducing Perlin loss $\mathcal{L}_{per}$—an auxiliary loss designed to stabilize early training. Rather than introducing a new loss function per se, Perlin loss leverages existing segmentation losses from $\mathcal{L}{cyc}$ but applies them to a novel set of synthetic images that closely resemble real fluorescence microscopy data. These images are generated using a 2D fractal Perlin noise texture~\cite{10.1145/325334.325247,Perlin}, which mimics chromatin structures, combined with background and foreground intensity variations derived from real microscopy observations, Gaussian blur, and Poisson noise. By enforcing self-supervision on these Perlin-generated images, we provide additional guidance to the segmentation network, particularly during early-stage training when the generator’s microscopy outputs may lack realism. See Figure~\ref{fig:main} for a visualization.

\noindent\textbf{Mask-to-Image Loss} When $G$ generates microscopy images for $S$, it may hallucinate or omit individual nuclei~\cite{Spcyclegan,utz2023focuscontentnoiseimproving,yao2022adganendtoendunsupervisednuclei}. We hypothesize that this occurs because such alterations do not influence the discriminator’s decision on whether an image appears real or fake. Since this hampers the training of $S$, we propose a novel loss function mask-to-image loss $\mathcal{L}_{m2i}$. The loss is composed of two sub losses $\mathcal{L}_\text{nuclei}$ to penalize the omission of nuclei and $\mathcal{L}_\text{bg}$ to penalize the addition of nuclei. The idea of $\mathcal{L}_\text{nuclei}$ is to enforce a minimum intensity for all pixels $\hat{c}_{i,j}$ in the generated image $\vec{c}$ where the corresponding mask pixels $m_{i,j} = 1$. It is defined as:
\begin{equation}
    \label{eq:m-to-i-loss-nuclei}
    \mathcal{L}_\text{nuclei}(\vec{m}_{i, j}, \hat{c}_{i, j}) = \mathbb{1}_{\vec{m}_{i, j} = 1} \cdot \max(t_n *(\hat{c}_\text{max} - \hat{c}_\text{min}) + \hat{c}_\text{min} - \hat{c}_{i, j}, 0),
\end{equation}
where $\hat{c}_\text{min}, \hat{c}_\text{max}$ are the minimum/maximum pixel intensities in the fake nuclei image and $t_n \in \left[0, 1 \right]$ is a relative threshold to control the influence of the loss. Similarly, we can define $\mathcal{L}_{bg}$ to prevent $G$ from producing nuclei where none exist in $\vec{m}$. We construct a dilated inverse mask $\overline{\vec{m}}$, which identifies background pixels with some flexibility near nuclei boundaries. The dilation expands each nucleus region outward by d pixels using a square kernel of size d, creating a transition zone where pixel classification remains unconstrained. This accounts for natural variations in nucleus morphology while ensuring that clear background regions remain free of artificial nuclei. Analogous to $\mathcal{L}_\text{nuclei}$, we can then define: 
\begin{equation}
    \label{eq:m-to-i-loss-background}
    \mathcal{L}_\text{bg}(\vec{m}_{i, j}, \hat{c}_{i, j}) = \mathbb{1}_{\overline{\vec{m}}_{i,j} = 1} \cdot \max(\hat{c}_{i, j} - \hat{c}_\text{min} -  t_b* (\hat{c}_\text{max} - \hat{c}_\text{min}), 0)
\end{equation}
where $t_b \in \left[0, 1 \right]$ is a relative threshold parameter to weight the loss. The mask-to-image loss combines both terms and is computed as 
\begin{equation}
    \label{eq:m-to-i-total}
    \mathcal{L}_\text{m2i}(\vec{m}, \hat{c}) = \frac{\lambda}{|\vec{m}|} \sum_{i, j} \left(\mathcal{L}_\text{nuclei}(\vec{m}_{i, j}, \hat{c}_{i, j}) + \mathcal{L}_\text{bg}(\vec{m}_{i, j}, \hat{c}_{i, j}) \right),
\end{equation}
where $\lambda$ scales the contribution to the total loss of CyclePose.

\noindent The final loss is then computed as sum of the individual weighted losses:
\begin{equation}
	\mathcal{L} = \mathcal{L}^S_{adv} + \mathcal{L}^G_{adv} + \mathcal{L}^G_{cyc} + \mathcal{L}^S_{cyc} + \mathcal{L}_{per} + \mathcal{L}_{m2i}   
\end{equation}
\section{Evaluation and Results}
\subsection{Datasets}
\label{sec:datasets}
We use two publicly available cell segmentation datasets from the Broad Biomage Benchmark Collection\footnote{\url{https://bbbc.broadinstitute.org/image_sets}}~\cite{ljosa_annotated_2012}. 

\noindent\textbf{U-2 OS (BBBC039)} 
This dataset by \cite{Caicedo2019-cu} consists of 200 fluorescence microscopy images of U-2 OS cells stained with Hoechst dye. It includes approximately 23,000 manually annotated nuclei. We used 100, 50, and 50 images for training, validation, and testing, respectively.

\noindent\textbf{DSB2018 (BBBC038)}
This diverse dataset was initially used in the 2018 Data Science Bowl~\cite{data-science-bowl-2018}. It contains nuclei images from various biological contexts. We used a curated subset of 497 images provided by~\cite{schmidt2018}, splitting it into 380 images for training, 67 for validation, and 50 for testing.

\noindent We apply random rotations, scaling, and translations to the images, then extract a 224 $\times$ 224 pixels crop centered on the transformed image. To synthesize masks, we randomly sample binary ellipses with major axis lengths chosen from the range $5$ to $30$ and eccentricities from $0.6$ to $0.9$, allowing up to $10\%$ overlap to simulate touching nuclei. To account for the non-ellipsoidal shape of real nuclei, we apply elastic deformation by generating a coarse displacement vector field of size $d \times d \times 2$ with independent Gaussian-distributed random variables $\mathcal{N}(0,\sigma^2)$, following \cite{Nisnet3d}. The displacement field is then smoothed via spline interpolation to match the image size. In our experiments, we sample $d$ from the range $5$ to $15$ and $\sigma^2$ from $1$ to $5$ per mask.

\noindent For model selection, we use a subset of 10 annotated image-mask pairs from the validation set, despite our segmentation approach being unsupervised. To maximize representativeness, we apply augmentation via 90-degree rotations and horizontal/vertical flipping. Supervised methods had access to the full validation set for model selection.
\subsection{Training}
We use the AdamW optimizer with a weight decay of $0.01$ and hyperparameters $\beta_1 = 0.5$ and $\beta_2 = 0.999$. The network is trained for $100$ epochs with an initial learning rate of $0.0008$, which is then linearly decayed to zero over an additional $100$ epochs. The batch size is set to $1$. Following the original CycleGAN implementation, we maintain an image pool of $50$ previously generated images to update the discriminator. The weighting of the image cycle loss is set to $10.0$, while the mask cycle loss is weighted at $15.0$. Similarly, we scale the Perlin segmentation loss by $15$, set the mask-to-image loss weight to $7.5$, and apply thresholds of $t_n = 0.2$ and $t_b = 0.3$ for nuclei and background, respectively. Hyperparameter selection was performed using the 10-image validation subset described in Section~\ref{sec:datasets}. Each experiment was repeated five times, and we report the mean and standard deviation across runs. Experiments were conducted on an NVIDIA A100 using PyTorch 2.5 and CUDA 12.4. 
\subsection{Compared Methods}
We compare CyclePose against both supervised and unsupervised baseline methods for nuclei segmentation, and report mean and std. dev. across five runs. %

\noindent\textbf{U-Net}
We train a supervised U-Net where the final layer predicts softmax logits for the two classes nuclei and background. We also evaluate a three-class variant with an additional boundary prediction. Instance segmentation masks are obtained by thresholding and extracting connected components.

\noindent\textbf{StarDist \& Cellpose} 
We train supervised StarDist~\cite{schmidt2018} and Cellpose~\cite{stringer_cellpose_2021} models using their respective default loss functions, learning rate schedules, and augmentation pipelines. The Cellpose architecture remains unchanged from its use in CyclePose. Additionally, we train an unsupervised variant of Cellpose using images generated with the Perlin noise-based synthesis approach described in Section~\ref{sec:methods}, along with their corresponding masks

\noindent\textbf{CycleGAN}
We use CycleGAN~\cite{CycleGAN2017} as an unsupervised baseline, replacing Cellpose with a conventional generator while maintaining the same network architecture as CyclePose. We evaluate CycleGAN in two ways: (1) as a direct segmentation model, applying a watershed algorithm to its binary mask output, and (2) as a synthetic data generator, where we first generate synthetic images and then train StarDist and Cellpose in a two-step approach, following \cite{utz2023focuscontentnoiseimproving}.

\noindent\textbf{DenoiSeg}
To compare our method with a (semi)-supervised approach that relies on a small set of annotated data, we trained a DenoiSeg~\cite{buchholz_denoiseg_2020} model using ten ground truth masks. 

\subsection{Evaluation Metrics}
Model performance was evaluated based on IoU between predicted and ground truth objects using the Jaccard index ($\operatorname{JAC}$). A predicted object is considered a true positive if its IoU with a ground truth object exceeds a predefined threshold ($\tau$). Predictions with IoU below $\tau$ are classified as false positives, while false negatives correspond to ground truth objects without a sufficiently overlapping prediction. The Jaccard index is calculated as $\operatorname{JAC} = \frac{TP}{TP+FP+FN}$.
In the nuclei segmentation literature, the Jaccard index has sometimes been mistakenly referred to as average precision (AP), which can cause confusion with the area under the precision recall curve~\cite{metrics2024reloaded}. To ensure clarity, we consistently refer to this metric as the Jaccard index. We report both $\operatorname{JAC}_{0.5}$, computed at $\tau = 0.5$, and the mean score $\operatorname{JAC}_{0.5:0.05:0.95}$, where $\tau$ is varied over the interval $[0.5, 0.95]$ in steps of 0.05. As a secondary evaluation metric for our ablation study, we use panoptic quality (PQ), which integrates detection performance and segmentation quality into a single score \cite{panoptic}. Segmentation quality is assessed by averaging the IoU scores of all true positive instances (mean matched score), while detection quality is captured by the $F_1$ score.
\subsection{Results}
\begin{figure}[tbp]
    \centering
    \includegraphics[width=\linewidth]{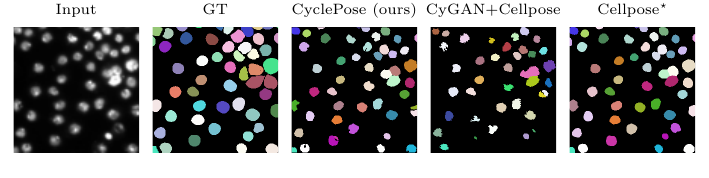}
    \caption{Qualitative comparison of CyclePose against top-performing unsupervised and supervised baseline methods. Methods trained with supervision are marked with $^\star$.}
    \label{fig:quali}
\end{figure}

\begin{table}[t]
    \centering
    \caption{Comparison of different models in terms of $\operatorname{JAC}$ (mean $\pm$ std.dev.) on U2OS and DSB2018 datasets. Best and second best results per category are highlighted in \textbf{bold} and \underline{underlined}. Perlin refers to images generated using masks + Perlin noise.}
    {\fontsize{8pt}{9pt}\selectfont
    \begin{tabular}{lcc@{\hskip 10pt}cc}
        \toprule
        \multirow{2}{*}{Method} & \multicolumn{2}{c}{U2--OS} & \multicolumn{2}{c}{DSB2018} \\
        \cmidrule(lr){2-3} \cmidrule(lr){4-5}
        & $\operatorname{JAC}_{0.5:0.05:0.95}$ & $\operatorname{JAC}_{0.5}$ & $\operatorname{JAC}_{0.5:0.05:0.95}$ & $\operatorname{JAC}_{0.5}$ \\
        \midrule
        \multicolumn{1}{l}{\textbf{Unsupervised}} \\
        \midrule
        CyclePose (ours) & \textbf{0.727} $\pm$ \textbf{0.018} & \textbf{0.887} $\pm$ \textbf{0.002} & \textbf{0.487} $\pm$ \textbf{0.008} & \textbf{0.764} $\pm$ \textbf{0.006} \\
        CyGAN & 0.479 $\pm$ 0.064 & 0.683 $\pm$ 0.047 & 0.289 $\pm$ 0.062 & 0.512 $\pm$ 0.109 \\
        CyGAN + StarDist & 0.495 $\pm$ 0.017 & 0.734 $\pm$ 0.015 & 0.336 $\pm$ 0.009 & \underline{0.651} $\pm$ \underline{0.005} \\
        CyGAN + Cellpose & 0.437 $\pm$ 0.007 & 0.632 $\pm$ 0.005 & \underline{0.413} $\pm$ \underline{0.003} & 0.645 $\pm$ 0.002 \\
        Perlin + Cellpose & \underline{0.655} $\pm$ \underline{0.009} & \underline{0.879} $\pm$ \underline{0.003} & 0.369 $\pm$ 0.022 & 0.636 $\pm$ 0.033 \\
        \midrule
        \multicolumn{1}{l}{\textbf{Supervised}} \\
        \midrule
        DenoiSeg (GT=10) & 0.572 $\pm$ 0.032 & 0.700 $\pm$ 0.030 & 0.403 $\pm$ 0.039 & 0.590 $\pm$ 0.037 \\
        2-class U-Net & 0.597 $\pm$ 0.004 & 0.721 $\pm$ 0.003 & 0.430 $\pm$ 0.012 & 0.638 $\pm$ 0.009 \\
        3-class U-Net & 0.701 $\pm$ 0.003 & 0.858 $\pm$ 0.003 & 0.493 $\pm$ 0.021 & 0.732 $\pm$ 0.024 \\
        StarDist & \underline{0.744} $\pm$ \underline{0.006} & \underline{0.904} $\pm$ \underline{0.003} & \underline{0.519} $\pm$ \underline{0.025} & \underline{0.815} $\pm$ \underline{0.013} \\
        Cellpose & \textbf{0.797} $\pm$ \textbf{0.002} & \textbf{0.918} $\pm$ \textbf{0.001} & \textbf{0.589} $\pm$ \textbf{0.019} & \textbf{0.839} $\pm$ \textbf{0.017} \\
        \bottomrule
    \end{tabular}
    }
    \label{tab:merged_results}
\end{table}
\begin{table}[t!]
\caption{Ablation study of loss functions in CyclePose, evaluated on DSB2018 using $\operatorname{JAC}$ and $\operatorname{PQ}$.}
\label{tab:ablation_losses}
\centering
{\fontsize{8pt}{9pt}\selectfont
\begin{tabular}{cccc@{\hskip 8pt}|cc@{\hskip 8pt}cc}
\toprule
$\mathcal{L}_{cyc}$ & $\mathcal{L}_{adv}$ & $\mathcal{L}_{per}$ & $\mathcal{L}_{m2i}$ & $\operatorname{JAC}_{0.5:0.05:0.95}$ & $\operatorname{JAC}_{0.5}$ & $\operatorname{PQ}_{0.5:0.05:0.95}$ & $\operatorname{PQ}_{0.5}$ \\
\midrule
\ding{51} & \ding{51} & \ding{51} & \ding{51} & 0.487 $\pm$ 0.008 & 0.764 $\pm$ 0.006 & 0.533 $\pm$ 0.008 & 0.719 $\pm$ 0.003 \\ %
\ding{51} & \ding{55} & \ding{51} & \ding{51} & 0.460 $\pm$ 0.007 & 0.742 $\pm$ 0.006 & 0.513 $\pm$ 0.006 & 0.701 $\pm$ 0.003 \\ %
\ding{51} & \ding{51} & \ding{55} & \ding{51} & 0.437 $\pm$ 0.022 & 0.725 $\pm$ 0.012 & 0.478 $\pm$ 0.025 & 0.681 $\pm$ 0.010 \\ %
\ding{51} & \ding{51} & \ding{51} & \ding{55} & 0.461 $\pm$ 0.046 & 0.743 $\pm$ 0.048 & 0.509 $\pm$ 0.044 & 0.700 $\pm$ 0.037 \\ %
\ding{51} & \ding{51} & \ding{55} & \ding{55} & 0.451 $\pm$ 0.025 & 0.742 $\pm$ 0.025 & 0.497 $\pm$ 0.022 & 0.694 $\pm$ 0.015 \\ %
\bottomrule
\end{tabular}
}
\end{table}
The performance of both supervised and unsupervised methods on U2-OS and DSB2018 is reported in Table~\ref{tab:merged_results}. Figure~\ref{fig:quali} provides a visual comparison of selected methods, highlighting their qualitative differences. Among unsupervised approaches, our method, CyclePose, achieves the best performance on both datasets by a substantial margin. On U2-OS, a simpler and more homogeneous dataset than DSB2018, the second-best unsupervised method is Cellpose trained on Perlin-stained masks, showing the effectiveness of this approach for unsupervised segmentation. Notably, despite being trained without annotated data, CyclePose surpasses the supervised 2-class U-Net on both datasets and even outperforms the 3-class U-Net on U2-OS. On the challenging DSB2018 dataset, CyclePose performs on par with the 3-class U-Net. However, a performance gap remains between CyclePose and fully-supervised Cellpose, which we consider a theoretical upper bound, given our method’s reliance on Cellpose as a backbone.

\noindent An ablation study was conducted on the DSB2018 dataset to assess the impact of different loss functions, including our proposed losses, with $\operatorname{JAC}$ and $\operatorname{PQ}$ reported in Table~\ref{tab:ablation_losses}. Networks trained without or solely with cycle loss failed to segment effectively and were excluded from results. Although omitting adversarial losses led to a slight performance drop, it offers the advantage of reduced computational cost by eliminating the need for two discriminator networks. Our proposed losses, $\mathcal{L}_{per}$ and $\mathcal{L}_{m2i}$, not only improve performance and enhance stability, as evidenced by the lower standard deviation of the results.
\section{Discussion}
In this work, we introduced CyclePose, an unsupervised nuclei segmentation method that leverages a cycle consistency loss within an adversarial training framework. Our approach enables high-quality nuclei segmentation without requiring annotated ground truth, making it particularly valuable in scenarios where annotations are hard to obtain.
By integrating Cellpose as the segmentation backbone, CyclePose benefits from a representation specifically optimized for cellular structures, leading to superior performance compared to methods that directly predict masks. However, despite eliminating the need for manual annotations, some expert knowledge remains necessary to configure the ellipse-based mask synthesis model used for training. Additionally, a small number of annotated images was needed for model selection.
A key limitation of our method is its reliance on synthetic masks generated from simple geometric primitives. This restricts its applicability to blob-like cellular structures such as nuclei. 
For future work, we aim to extend CyclePose to 3-D volumetric segmentation, where annotation costs are even more prohibitive. Additionally, exploring more sophisticated shape priors beyond simple ellipses could broaden the applicability of our framework to diverse cell types and imaging conditions.

\begin{credits}
\subsubsection{\ackname} J.\@ U., S.\@ U., and K.\@ B. received funding by Deutsche Forschungsgemeinschaft (DFG, German Research Foundation) - 405969122. The authors gratefully acknowledge the scientific support and HPC resources provided by the Erlangen National High Performance Computing Center (NHR@FAU) of the Friedrich-Alexander-Universität Erlangen-Nürnberg (FAU) under the NHR project b221cb. NHR funding is provided by federal and Bavarian state authorities. NHR@FAU hardware is partially funded by the German Research Foundation (DFG) – 440719683. 

\end{credits}

\bibliographystyle{splncs04}
\bibliography{lit, lit_jonas}
\end{document}